\documentclass[conference]{IEEEtran}
\IEEEoverridecommandlockouts
\usepackage{cite}
\usepackage{amsmath,amssymb,amsfonts}
\usepackage{graphicx}
\usepackage{textcomp}
\usepackage{xcolor}
\usepackage{pifont}
\usepackage{bbding}
\usepackage{authblk}
\usepackage{booktabs}
\usepackage{tabularx}
\usepackage{makecell}
\usepackage{algorithm,algorithmic}
\usepackage{booktabs}

\usepackage{multirow} 
\usepackage{hyperref}
\hypersetup{
    colorlinks=true,
    linkcolor=blue, 
    citecolor=blue, 
    urlcolor=black 
}
\usepackage[capitalize]{cleveref}
\usepackage{adjustbox}

\usepackage{xspace}
\makeatletter
\DeclareRobustCommand\onedot{\futurelet\@let@token\@onedot}
\def\@onedot{\ifx\@let@token.\else.\null\fi\xspace}

\makeatother

\def\BibTeX{{\rm B\kern-.05em{\sc i\kern-.025em b}\kern-.08em
    T\kern-.1667em\lower.7ex\hbox{E}\kern-.125emX}}
\begin{document}

\title{
DTEA: Dynamic Topology Weaving and Instability-Driven Entropic Attenuation for Medical Image Segmentation
}

\author[1]{Weixuan Li}
\author[1]{Quanjun Li}
\author[2,3]{Guang Yu}
\author[1]{Song Yang}
\author[2*]{Zimeng Li\thanks{* Corresponding authors: li\_zimeng@szpu.edu.cn, xuhangc@hzu.edu.cn}}
\author[4]{\authorcr Chi-Man Pun}
\author[5,6]{Yupeng Liu}
\author[4,7*]{Xuhang Chen\thanks{
This work was supported in part by Shenzhen Medical Research Fund (Grant No. A2503006), in part by the National Natural Science Foundation of China (Grant No. 62501412 and 82300277), in part by Shenzhen Polytechnic University Research Fund (Grant No. 6025310023K), in part by Medical Scientific Research Foundation of Guangdong Province (Grant No. B2025610 and B2023012), in part by the Science and Technology Development Fund, Macau SAR, under Grant 0193/2023/RIA3 and 0079/2025/AFJ, and the University of Macau under Grant MYRG-GRG2024-00065-FST-UMDF, and in part by Guangdong Basic and Applied Basic Research Foundation (Grant No. 2024A1515140010).}}

\affil[1]{School of Advanced Manufacturing, Guangdong University of Technology}
\affil[2]{School of Electronic and Communication Engineering, Shenzhen Polytechnic University}
\affil[3]{Guangzhou City University of Technology}
\affil[4]{Faculty of Science and Technology, University of Macau}
\affil[5]{Department of Cardiology, Guangdong Provincial People's Hospital (Guangdong Academy of Medical Sciences),\protect\\Southern Medical University}
\affil[6]{Guangdong Cardiovascular Institute, Guangdong Provincial People's Hospital,\protect\\Guangdong Academy of Medical Sciences}
\affil[7]{School of Computer Science and Engineering, Huizhou University}

\maketitle

\begin{abstract}

In medical image segmentation, skip connections are used to merge global context and reduce the semantic gap between encoder and decoder. Current methods often struggle with limited structural representation and insufficient contextual modeling, affecting generalization in complex clinical scenarios. We propose the DTEA model, featuring a new skip connection framework with the Semantic Topology Reconfiguration (STR) and Entropic Perturbation Gating (EPG) modules. STR reorganizes multi-scale semantic features into a dynamic hypergraph to better model cross-resolution anatomical dependencies, enhancing structural and semantic representation. EPG assesses channel stability after perturbation and filters high-entropy channels to emphasize clinically important regions and improve spatial attention. Extensive experiments on three benchmark datasets show our framework achieves superior segmentation accuracy and better generalization across various clinical settings. The code is available at \href{https://github.com/LWX-Research/DTEA}{https://github.com/LWX-Research/DTEA}.

\end{abstract}


\section{INTRODUCTION}
Medical image segmentation plays a vital role in clinical diagnosis and treatment planning~\cite{1coates2015tailoring}, yet practical deployment remains challenging due to image noise, signal heterogeneity, and structural complexity~\cite{2riccio2018new}, all of which severely hinder model generalization. Additionally, large inter-patient anatomical variability, limited annotated data, and modality-specific artifacts further constrain the applicability of natural image segmentation methods in medical domains~\cite{3litjens2017survey}. These issues highlight the need for more robust and adaptable segmentation strategies to ensure reliable performance in complex clinical environments.

In the  medical image segmentation domain, the U-shaped architecture has emerged as the mainstream paradigm. It typically consists of an encoder, a decoder, and skip connections. Models represented by UNet~\cite{6ronneberger2015u} have demonstrated strong  performance across various tasks but still struggle to effectively capture complex anatomical structures while maintaining semantic consistency. Current medical image segmentation networks predominantly adopt an encoder-decoder architecture, often based on convolutional neural networks (CNNs), to extract hierarchical features and reconstruct fine-grained segmentation maps~\cite{21xu2023dcsau, 22srivastava2021msrf,liu2023explicit,zhu2024test}. CNNs are well-suited for capturing local structures~\cite{li2023cee,liu2023coordfill,li2022monocular,li2022few,li2024cross,liu2024dh,lei2025cmamrnet}, but their limited receptive fields constrain the ability to model long-range dependencies and global anatomical contexts~\cite{7hatamizadeh2022unetr}. To overcome these limitations, Transformer-based architectures have been introduced~\cite{li2025adaptive}, leveraging self-attention mechanisms to strengthen global semantic modeling and representation~\cite{8chen2024transunet, 9cao2022swin,liu2024depth,zheng2024smaformer}. However, despite these advances, performance bottlenecks remain, particularly in bridging the semantic gap between the encoder and decoder due to insufficient information transfer~\cite{10zhang2021transfuse}.

To alleviate this issue, skip connections have been widely employed to facilitate feature fusion across different semantic levels~\cite{11ates2023dual}. Early designs like UNet~\cite{6ronneberger2015u} use direct connections to integrate low-level and high-level features, while more advanced variants such as UNet++~\cite{12zhou2018unet++} introduce dense skip pathways to enhance multi-scale fusion. Recent Transformer-based methods like CFATransUNet~\cite{13wang2024cfatransunet} further refine skip connections by leveraging global context modeling. 
Despite these advances, these methods still suffer from attention instability and susceptibility to background noise. Therefore, improving the design of skip connections to enhance cross-scale semantic consistency and suppress attention ambiguity remains an open and critical direction for advancing medical image segmentation.
\begin{figure}[ht]
\centerline{\includegraphics[width=\linewidth]{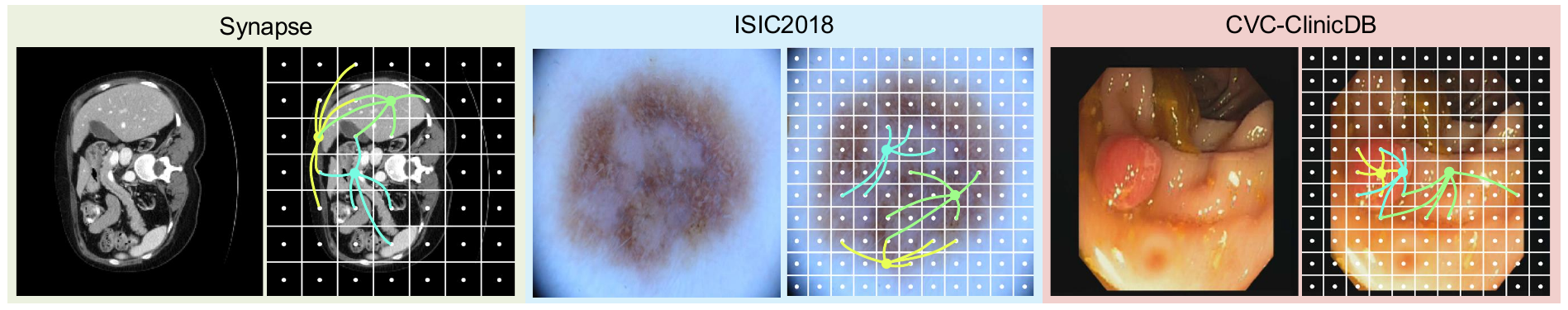}}
\caption{Hypergraph visualization of DTEA. Three patches (Yellow, Blue, and Green) are selected as central nodes to visualize the corresponding hyperedges generated by the STR module. The lesion and non-lesion areas exhibit a clear separation in the hypergraph. Within the lesion region, the hyperedges show strong aggregation, while nodes in the boundary region also display notable similarity and structural correlation.}
\label{fig:hypergraph}
\end{figure}

To address the semantic gap between the encoder and decoder and effectively capture both local and global dependencies in complex visual tasks, this paper proposes a Dynamic Topology Weaving and Instability-Driven Entropic Attenuation for Medical Image Segmentation (DTEA) model. 
Specifically, the model uses Transformers as both encoder and decoder to extract long-range and local semantic features, efficiently transferring information through skip connections that integrate the Semantic Topology Reconfiguration (STR) and Entropic Perturbation Gating (EPG).
STR dynamically constructs a hypergraph representing cross-scale anatomical structures, flexibly capturing structural dependencies at different resolutions and adaptively enhancing key semantic representations. EPG amplifies stability differences among channels through nonlinear chaotic mapping and uses entropy as an uncertainty measure to dynamically suppress channels with high ambiguity and low information content, considerably improving the discriminability and focus of spatial attention. This innovative skip connection mechanism is compatible with various backbone architectures, including CNNs and Transformers, substantially enhancing segmentation performance and improving the visual separability of lesion regions, as illustrated in \cref{fig:hypergraph}. Extensive experimental results demonstrate that the proposed modules outperform existing methods across multiple medical image segmentation tasks, exhibiting strong robustness, efficiency, and generalization capability.
The contributions of this paper are summarized as follows:
\begin{enumerate}
    \item We propose DTEA for medical image segmentation, leveraging the Transformer backbone and an innovative skip connection framework that integrates the STR and EPG modules to bridge the encoder-decoder semantic gap and enhance multi-scale feature fusion.
    \item STR dynamically constructs multi-scale features into a hypergraph structure, enabling explicit modeling of cross-resolution high-order anatomical dependencies and enhancing semantic consistency in feature fusion.
    \item EPG integrates nonlinear chaotic perturbation with entropy-driven channel selection to suppress information redundancy and ambiguous attention, thereby improving the discriminability and focus of spatial attention.
\end{enumerate}


\begin{figure}[ht]
\centerline{\includegraphics[width=\linewidth]{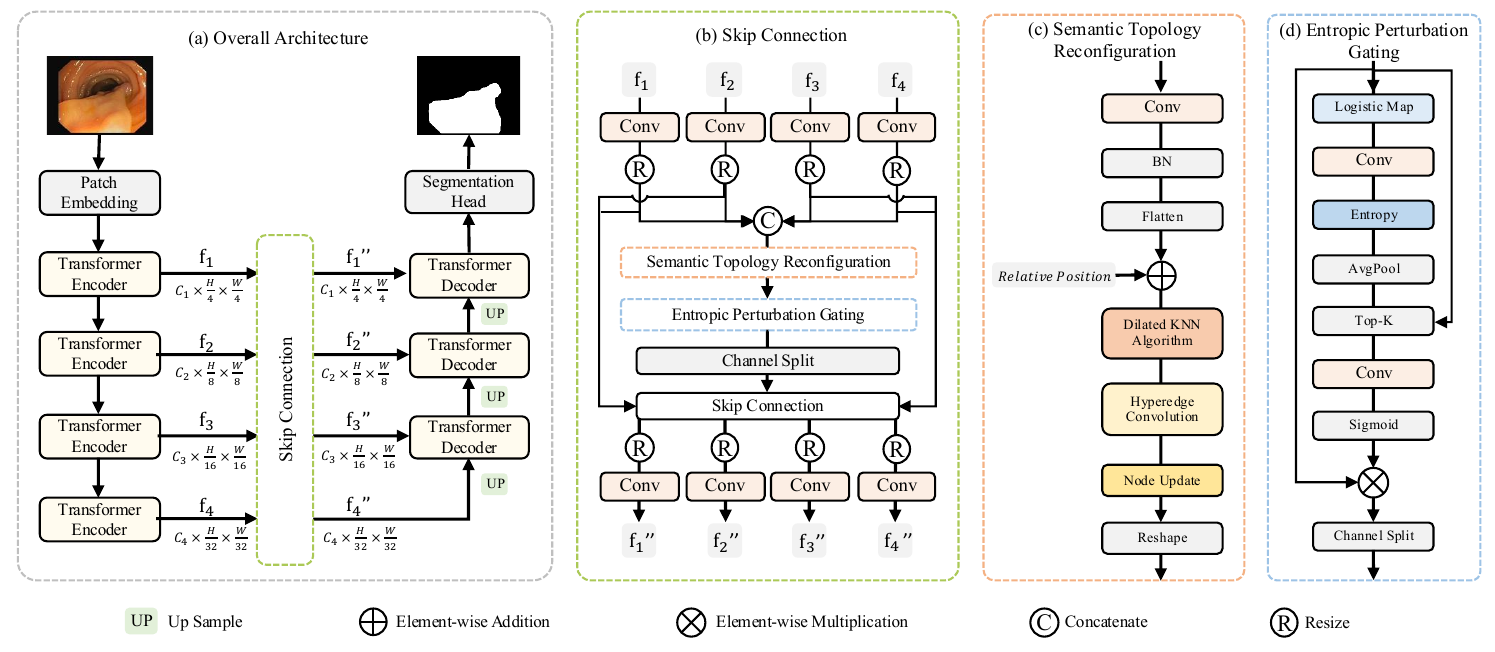}}
\caption{(a) The overall architecture of the proposed DTEA. (b) The skip connection framework. (c) Semantic Topology Reconfiguration (STR). (d) Entropy Perturbation Gating (EPG).}
\label{fig:overall}
\end{figure}

\section{METHOD}
\Cref{fig:overall} illustrates the overall architecture of the proposed DTEA, which adopts a U-shaped network architecture. We incorporate Transformer blocks as the core components of both the encoder and decoder. In addition, we introduce a novel skip connection mechanism composed of four stages: (i) Feature Preprocessing; (ii) Semantic Topology Reconfiguration (STR); (iii) Entropic Perturbation Gating (EPG); and (iv) Feature Postprocessing.

\subsection{Feature Preprocessing}
We utilize the feature maps outputted from the four stages of the encoder, denoted as $f_i \in \mathbb{R}^{C_i \times \frac{H}{2^{i+1}} \times \frac{W}{2^{i+1}}} $ for $ i = 1, 2, 3, 4 $, as inputs to the skip connections, where $ C_i $ represents the number of channels and $(H, W)$ denote the spatial dimensions of the input image. Since these feature maps differ in both spatial resolution and channel dimension, a unified transformation is applied prior to fusion.  Specifically, to reduce the decoder's computational burden and ensure spatial alignment, each feature map is first passed through a $ 1 \times 1 $ convolution to compress the channel dimension to a fixed size $ C_s=32 $. The resulting features are then resized to a target resolution $ H_t = \frac{H}{32}, W_t = \frac{W}{32} $, which corresponds to the output resolution of the fourth encoder stage. This process can be formulated as:
\begin{equation}
f_i' = \mathrm{Resize}_{(H_t, W_t)} \Big( \mathrm{Conv}(f_i) \Big) \in \mathbb{R}^{C_s \times H_t \times W_t},
\end{equation}
where $\mathrm{Conv}$ denotes the 2D convolution operation and $ \mathrm{Resize}_{(H_t, W_t)} $ denotes spatial resizing to the target resolution. Subsequently, the four feature maps are concatenated along channels to generate a multi-scale representation:
\begin{equation}
    f_{concat} = \mathrm{Concat}(f_1', f_2', f_3', f_4') \in \mathbb{R}^{C \times H_t \times W_t},
\end{equation}
where $ C = 4C_s $ is the aggregated channel dimension resulting from concatenation.

\subsection{Semantic Topology Reconfiguration}
To effectively capture complex semantic dependencies beyond simple pairwise relationships, we model the feature interactions using a hypergraph structure. In a hypergraph, a hyperedge links a central node with multiple spatially-aware neighbors simultaneously, enabling the representation of higher-order semantic relationships and enriching the model capacity to capture non-local contextual information~\cite{34han2023vision}.

STR first refines the multi-scale feature map $ f_{concat} \in \mathbb{R}^{C \times H_t \times W_t} $ using convolutional layers followed by normalization, then reshapes the refined feature map into a node matrix.
We explicitly incorporate relative position encoding into the node features to enhance spatial awareness. A dilated K-nearest neighbor algorithm is then applied to construct a sparse adjacency relation, based on which hyperedges are formed. Each hyperedge $e$ consists of a center node $x$ and its adjacent nodes. To capture the complex semantic dependencies among nodes within lesion regions, we design a novel hyperedge convolution operation to aggregate features within each hyperedge, defined as:
\begin{equation}
h_e = {x} + \sum_{x_j \in e} \sigma(\alpha c_j + \beta) \cdot x_j,
\end{equation}
where $ \sigma $, $ \alpha $ and $ \beta $ denote the sigmoid activation function and two learnable scalars, and $ c_j $ denotes the cosine similarity between central node $ x $ and its neighbor $ x_j $. To feed semantic information back to the nodes, we collect the aggregated features from their associated hyperedges and update the node representations via a reverse information flow, with the aggregation function defined as follows:
\begin{equation}
x' = \sigma\bigg( \mathrm{Conv}\Big(
(1 + \varepsilon) x + \sum_{e \in {N}} h_e
\Big) \bigg),
\end{equation}
where $ {N} $ is the set of hyperedges containing node $ x $, and $ \varepsilon $ is a modulation factor. This hierarchical formulation bridges fine-grained node-level features with higher-order relational context in a unified and efficient message-passing framework. After updating the node features, the node matrix is reshaped back into the spatial feature map $f_{STR} \in \mathbb{R}^{C \times H_t \times W_t}$ to restore the original spatial structure. 

\begin{figure}[t]
\centerline{\includegraphics[width=\columnwidth]{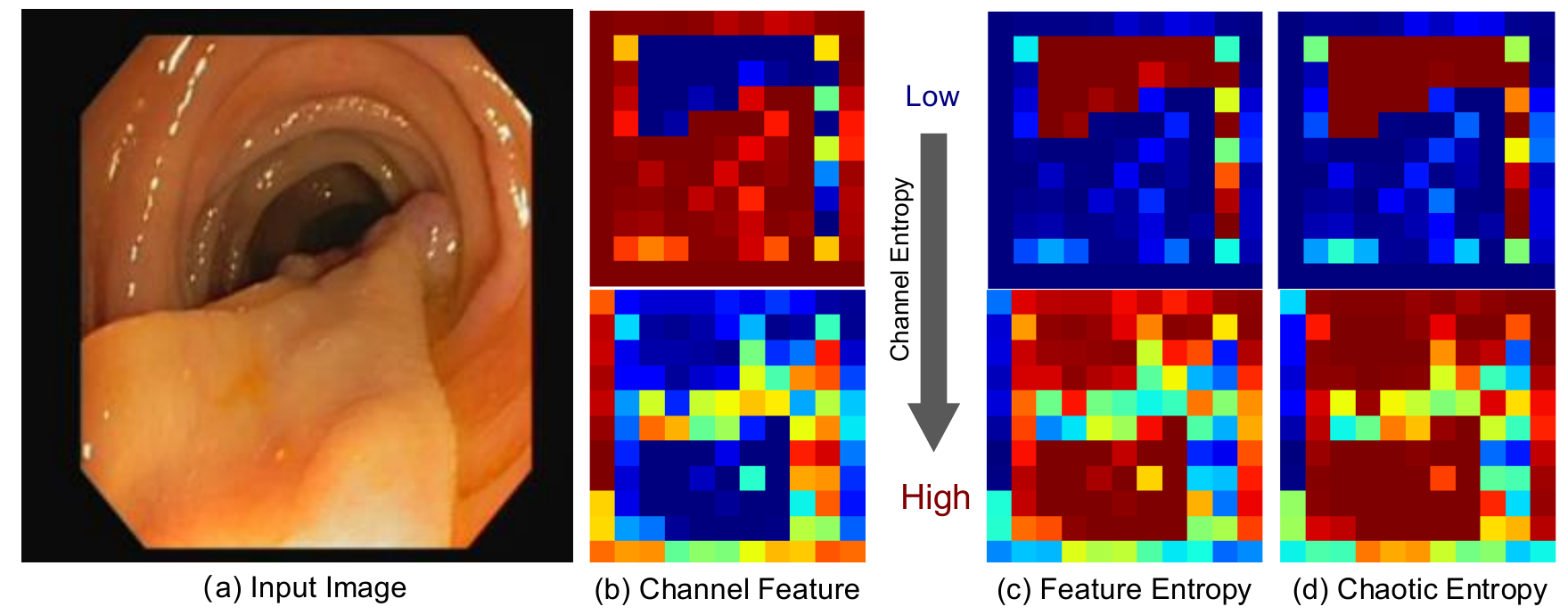}}
\caption{Visual comparison of low-entropy and high-entropy channels in EPG. (a) Input image. (b) Channels Feature maps. (c) The entropy maps of the feature maps. (d) The entropy maps of the feature maps after chaotic perturbation.}
\label{fig:entropy}
\end{figure}
\subsection{Entropic Perturbation Gating}
Although STR can capture global dependencies, noise and complex structures in medical images still lead to high-entropy, non-informative channels, which degrade the quality of spatial attention maps~\cite{19chen2021lesion}. To address this, we propose EPG that suppresses channels with high information entropy after chaotic perturbation to enhance feature representation, as illustrated in \cref{fig:entropy}. Specifically, we introduce chaotic perturbations based on the Logistic Map~\cite{20may1976simple} to probe the intrinsic stability of each input feature channel, and then apply convolution to the perturbed features to aggregate neighborhood information. The formulation is as follows:
\begin{equation}
f_{chaotic} = \mathrm{Conv} \bigl(f_{STR} \cdot \mu \cdot (1 - f_{STR}) \bigr),
\end{equation}
where $\mu$ is the chaos coefficient controlling the perturbation strength. To enhance the perturbation effect while preserving the sensitivity of chaotic dynamics, we set $\mu$ to 3.99. Semantic channels remain stable under perturbation, whereas unstructured channels quickly lose correlation and tend toward a high-entropy state.

To evaluate the stability and information complexity of each channel under perturbation, we introduce an entropy-based gating mechanism based on Shannon entropy, computing pixel-wise spatial uncertainty for each channel. The entropy score $ E \in \mathbb{R}^{C} $ is defined as:
\begin{equation}
E = \mathbb{E}_{h,w}(P \log P),
\end{equation}
where $ \mathbb{E}_{h,w} $ represents the mean expectation over the spatial dimensions $ (h, w) $, and $ P = \sigma(f_{chaotic}) $.
Building upon these entropy values, we then perform channel pruning by selecting a sparse subset of the $K$ channels with the lowest entropy to generate spatial attention as follows:
\begin{equation}
    f_{EPG} = f_{STR} \cdot \sigma
    \bigg(
        \mathrm{Conv}
        \Big(
            \operatorname{Top-k}(f_{STR}, -E, K)
        \Big)
    \bigg),
\end{equation}
where $\operatorname{Top-k}$ denotes selecting the feature subset corresponding to the $K$ channels with the lowest entropy $E$. This entropy-guided channel selection, combined with global dependency modeling, results in spatial attention maps with improved stability and discriminative power.

\begin{table}[ht]
    \centering
    \caption{Results of comparison experiments on Synapse Dataset. And DSC for each organ class are reported.}
    \adjustbox{width=\linewidth}{
    \begin{tabular}{ccccccccccc}
        \toprule
        \textbf{Model}& \textbf{DSC}& \textbf{HD95}& \textbf{Aorta} & \textbf{Gallbladder} & \textbf{Kidney(L)}& \textbf{Kidney(R)}& \textbf{Liver} & \textbf{Pancreas} & \textbf{Spleen} & \textbf{Stomach} \\
        \midrule
        UNet\cite{6ronneberger2015u}    & 70.1& 40.1& 84.3 & 44.6&  73.3&  72.3& 92.0& 47.0& 79.2 & 68.0 \\
        DCSAU-Net\cite{21xu2023dcsau}  & 72.0&38.6& 82.4& 53.9&  77.1&  69.8& 92.7 & 47.4& 84.6 & 68.0 \\
        MSRF-Net\cite{22srivastava2021msrf}  & 76.9&36.2& 85.6& 58.7&  82.8&  73.6& 94.6 & 57.3 & 88.3 & 75.4 \\
        TransUNet\cite{8chen2024transunet}  & 77.5 & 34.3 & 87.2 & 63.2&  81.9&  77.0& 94.1 & 55.9 & 85.1 & 75.6 \\
        SwinUnet\cite{9cao2022swin}& 78.7& 19.5& 85.3& 67.4& 84.8& 81.3& 94.2& 55.2& 88.6&72.7\\
        G-CASCADE\cite{17rahman2024g}& 78.2& 18.8& 85.1 & 58.6&  82.2&  79.3& 94.9 & 55.2 & 88.7 & 82.7 \\
        M$^2$SNet\cite{15zhao2023m}  & 77.5&24.4& 84.9 & 56.6&  79.2&  76.1& 94.9 & 58.4& 89.1 & 80.5\\
        UNet++\cite{12zhou2018unet++}  & 78.2& 34.3& 85.4& 64.8&  80.7&  77.0& 93.4& 59.9& 88.7& 80.1\\
        MADGNet\cite{23nam2024modality}  & 80.6 &24.9& 86.0& 66.5&   83.9&  79.3& 94.8& 63.6& 90.2& 80.5\\
        CFATransUnet\cite{13wang2024cfatransunet}& 81.8&23.8& 85.5& 66.8&  85.8&  81.8& 94.0& 64.9& 91.1& 84.5\\
        \midrule
        \textbf{DTEA}& \textbf{83.2}& \textbf{14.5} & \textbf{86.9} & \textbf{67.8}& \textbf{86.7}& \textbf{83.1}& \textbf{95.4}& \textbf{68.1}& \textbf{91.5}& \textbf{85.5}\\
        \bottomrule
    \end{tabular}}
    \label{tab:synapse}
\end{table}

\begin{table}[t]
\centering
\caption{Comparison experiments on ISIC 2018 and CVC-ClinicDB Dataset.}
\label{tab:isic2018_cvc}
\begin{tabular}{@{}ccccc@{}}
\toprule
\multirow{2}{*}{\textbf{Model}} & \multicolumn{2}{c}{\textbf{ISIC2018}} & \multicolumn{2}{c}{\textbf{CVC-ClinicDB}} \\
\cmidrule(r){2-3} \cmidrule(r){4-5}
& \textbf{DSC}& \textbf{mIoU}& \textbf{DSC}& \textbf{mIoU}\\
\midrule
UNet\cite{6ronneberger2015u}     & 86.7 & 79.1 & 76.9& 69.1 \\
DCSAU-Net\cite{21xu2023dcsau} & 89.0 & 82.0 & 80.6& 73.7\\
MSRF-Net\cite{22srivastava2021msrf}  & 88.2 & 81.3 & 83.2& 76.5\\
TransUNet\cite{8chen2024transunet}& 87.3& 81.2& 90.5& 84.7\\
SwinUnet\cite{9cao2022swin} & 86.7& 78.4& 83.8& 75.3\\
G-CASCADE\cite{17rahman2024g} & 90.4& 84.2& 92.0& 87.6\\
M$^2$SNet\cite{15zhao2023m}   & 89.2 & 83.4 & 91.9& 87.7\\
UNet++\cite{12zhou2018unet++}   & 87.3 & 80.2 & 82.3& 75.8\\
MADGNet\cite{23nam2024modality}   & 90.2& 83.7 & 92.6& 88.0\\
CFATransUnet\cite{13wang2024cfatransunet} & 90.3 & 83.6 & 91.0& 86.2\\
\midrule
\textbf{DTEA}& \textbf{91.9} & \textbf{85.8} & \textbf{93.4}& \textbf{88.7}\\
\bottomrule
\end{tabular}
\end{table}
\subsection{Feature Postprocessing}
The fused feature ${f}_{EPG}$ is evenly split along the channel dimension into four sub-features, each containing $C_s$ channels. Each sub-feature is then added to the corresponding encoder feature $f_i’$ via a residual connection to enhance feature stability and information flow. The fused result is first resized to match the spatial resolution required by the decoder, and then further restored to a higher spatial dimension through a convolutional layer, formulated as:
\begin{equation}
{f}_{i}^{\prime\prime} =
\mathrm{Conv} \left(
\mathrm{Resize}_{\left(\frac{H}{2^{i+1}}, \frac{W}{2^{i+1}}\right)}
\left( f_i' + {f_{EPG}}_i \right)
\right).
\end{equation}

\section{EXPERIMENT RESULTS}
\subsection{Dataset}
To verify the robustness and generalizability of the proposed approach, we evaluate it across three public datasets encompassing diverse tasks, including multi-organ, skin lesion, and polyp segmentation.

\textbf{Synapse}: Synapse~\cite{24landman2015miccai} is a publicly available multi-organ segmentation dataset comprising 30 abdominal CT scans, with a total of 3,779 axial slices. The images are annotated for eight abdominal organs: aorta, gallbladder, left kidney, right kidney, liver, pancreas, spleen, and stomach. Following previous studies~\cite{9cao2022swin,13wang2024cfatransunet}, we use 18 cases for training and the remaining 12 for testing. 

\textbf{ISIC 2018}: The ISIC 2018~\cite{25codella2019skin} dataset, released by the International Skin Imaging Collaboration, focuses on lesion segmentation from dermoscopic images. It contains a total of 2,594 images with varying resolutions. Following previous studies~\cite{23nam2024modality}, we randomly split the dataset into 1,868 training images, 465 validation images, and 261 testing images.

\textbf{CVC-ClinicDB}: The CVC-ClinicDB~\cite{26bernal2015wm} dataset is collected from 23 standard white-light colonoscopy video sequences, consisting of 612 colonoscopic images annotated with corresponding lesion segmentation masks. Following the data split used in prior studies~\cite{23nam2024modality}, we use 428 images for training, 61 for validation, and 123 for testing.

\subsection{Implementation Details}
We set the batch size to 24 and adopted the AdamW optimizer with an initial learning rate of 3e-4. The learning rate was adjusted using the CosineAnnealingLR scheduler, with a minimum value of 6e-7. To improve model generalization, we also applied data augmentation techniques such as random flipping and random rotation. For the ISIC 2018 and CVC-ClinicDB datasets, the input resolution was set to $352 \times 352$, and the model was trained for 100 epochs using the BceDice loss. We provide detailed evaluations using multiple metrics, including mean Intersection over Union (mIoU) and Dice Similarity Coefficient (DSC). For the Synapse dataset, we used an input resolution of $224 \times 224$ and trained the model for 150 epochs with the CeDice loss. We report DSC scores for individual organs, along with the 95th percentile Hausdorff Distance (HD95) between the predicted and ground truth segmentations.
\begin{figure*}[ht]
\centerline{\includegraphics[width=\textwidth]{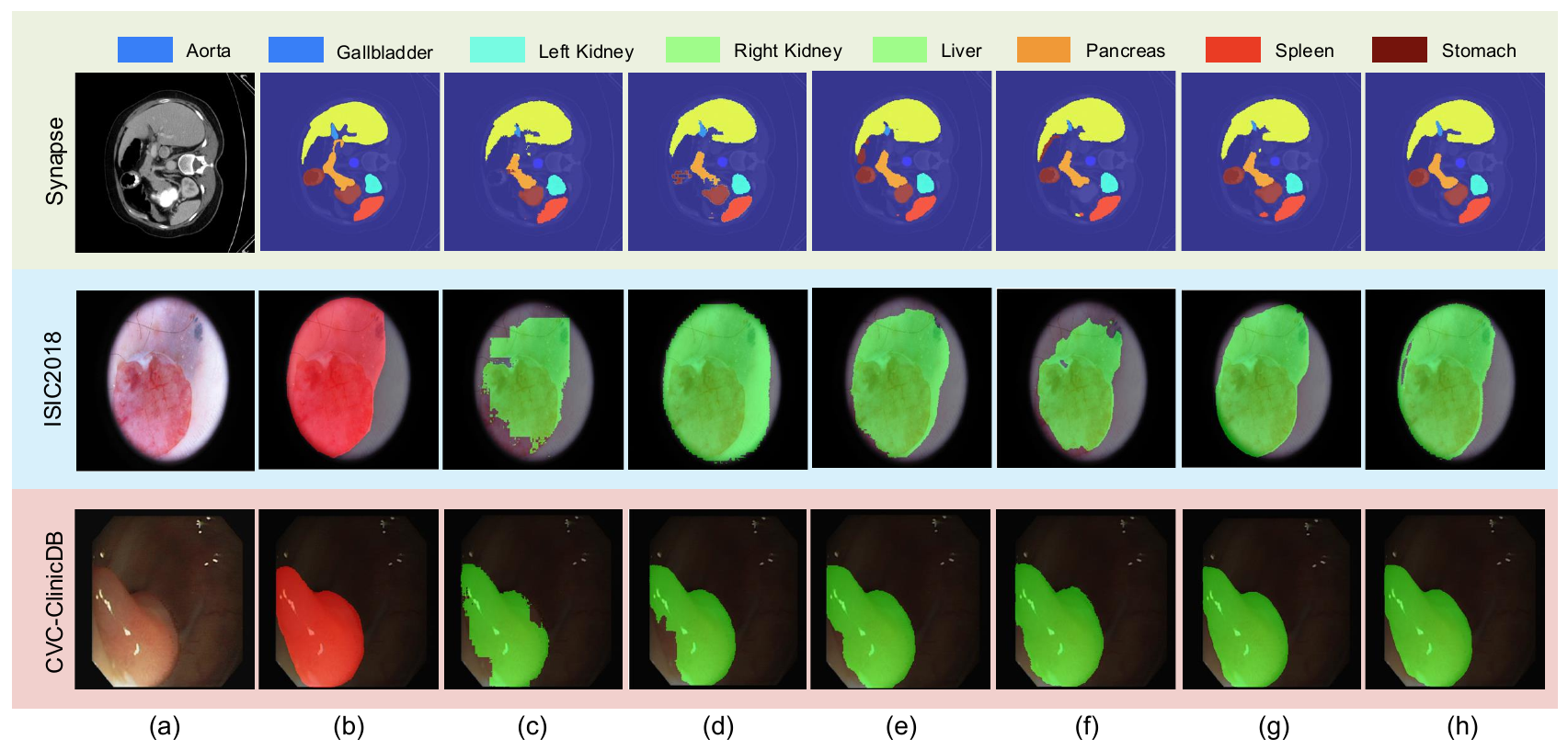}}
\caption{Visual Comparison for other model and proposed DTEA. (a) Input image. (b) Ground Truth. (c) Unet. (d) TransUNet. (e) M$^2$SNet. (f) MADGNet. (g) CFATransUnet. (h) DTEA.}
\label{fig:comparison}
\end{figure*}
\subsection{Comparison with State-of-the-art Models}
We conducted a comprehensive evaluation of the proposed DTEA on three public datasets. To validate its effectiveness, we compared DTEA with a diverse set of state-of-the-art methods, including traditional convolutional networks such as UNet~\cite{6ronneberger2015u}, DCSAU-Net~\cite{21xu2023dcsau}, and MSRF-Net~\cite{22srivastava2021msrf}; Transformer-based approaches such as TransUNet~\cite{8chen2024transunet} and Swin-Unet~\cite{9cao2022swin}; hybrid architectures incorporating graph neural networks such as G-CASCADE~\cite{17rahman2024g}; and multi-scale fusion frameworks such as M$^2$SNet~\cite{15zhao2023m}, UNet++~\cite{12zhou2018unet++}, MADGNet~\cite{23nam2024modality}, and CFATransUNet~\cite{13wang2024cfatransunet}. 

The quantitative results, presented in ~\cref{tab:synapse} and~\cref{tab:isic2018_cvc}, demonstrate that DTEA consistently achieves leading performance across all tasks. On the Synapse dataset, DTEA achieved a Dice score of 83.2\% and an HD95 of 14.5 mm, outperforming all competing models. Furthermore, compared to the recent CFATransUNet, DTEA achieved additional Dice score improvements of 1.6\% and 2.4\% on the ISIC 2018 and CVC-ClinicDB datasets, respectively, further validating its generalizability and robustness across different segmentation tasks.

As shown in ~\cref{fig:comparison}, the qualitative results across datasets further validate the robustness of DTEA. By dynamically integrating multi-scale features through a topological structure, DTEA accurately localizes lesions and preserves clear boundaries even in the presence of complex structures or blurry edges, demonstrating the adaptability and potential of its skip connection design in medical image segmentation.

\begin{table}[t]
\centering
\caption{Ablation experiments of STR and EPG modules validity on Synapse dataset. The baseline only use a direct skip connection in DTEA.}
\label{tab:Ablation_STR_EPG}
  \begin{tabular}{ccccc}
    \toprule
    \textbf{Baseline} & \textbf{STR} & \textbf{EPG} & \textbf{DSC}& \textbf{HD95} \\
    \midrule
    $\checkmark$ & $\times$     & $\times$      & 80.0 & 26.7 \\
    $\checkmark$ & $\times$     & $\checkmark$  & 82.8 & 20.3\\
    $\checkmark$ & $\checkmark$ & $\times$      & 82.7 & 21.1\\
    $\checkmark$ & $\checkmark$ & $\checkmark$  & \textbf{83.2}& \textbf{14.5}\\
    \bottomrule
  \end{tabular}
\end{table}

\subsection{Ablation Study}
\subsubsection{Effectiveness analysis of STR and EPG}
To evaluate the effectiveness of STR and EPG, we conducted ablation studies on the Synapse dataset, as shown in \cref{tab:Ablation_STR_EPG}. Results demonstrate that integrating either module individually leads to significant performance improvements. STR leverages a multi-scale hypergraph structure to capture high-order anatomical relationships across resolutions, avoiding redundant connections introduced by conventional adjacency methods. EPG employs channel-wise entropy evaluation to filter noisy channels, enhancing the focus of the spatial attention mechanism and suppressing background interference. The synergy of these two modules ensures that the features transmitted through skip connections are both semantically rich and low-noise, which is essential for accurate segmentation under challenging clinical conditions.

\begin{table}[t]
\centering
\caption{Ablation Study on $K$-value in EPG}
\label{tab:k}
  \begin{tabular}{ccc}
    \toprule
    \textbf{$K$} & \textbf{DSC} & \textbf{HD95} \\
    \midrule
    16  &  79.4& 35.1\\
    32  &82.1  & 18.4 \\
    \textbf{64} &\textbf{83.2}& \textbf{14.5}\\
    128  &82.2 &22.4 \\
    \bottomrule
  \end{tabular}
\end{table}
\subsubsection{$K$-value in EPG}
\Cref{tab:k} shows the performance of our model on the Synapse multi-organ segmentation task under different channel selection numbers $K \in {16, 32, 64, 128}$. The results indicate that when $K=64$, meaning half of the total channels $4C_s=128$ are selected, the model achieves the best performance in both DSC and HD95 metrics. This suggests that appropriate channel selection can reduce redundancy while preserving key features, thereby enhancing spatial attention mechanisms and overall performance. When using all channels with $K = 128$, performance decreases, indicating that excessive redundant features may negatively impact spatial modeling. These findings highlight that proper channel selection is crucial for improving spatial representation capability and enhancing model robustness.

\begin{table}[t]
\centering
\caption{Ablation Study on DTEA with different backbone.}
\label{tab:backbone_grouped}
\begin{tabular}{cccc}
  \toprule
  \textbf{Network Type} & \textbf{Backbone} & \textbf{DSC} & \textbf{HD95} \\
  \midrule
  \multirow{3}{*}{CNN} & ResNet-50 & 80.4 & 24.8 \\
                       & Res2Net   & 82.4 & 21.3 \\
                       & ResNeSt-50& 79.3 & 31.5 \\
  \midrule
  \multirow{3}{*}{Transformer} & ViT-B/16  & 74.2 & 28.7 \\
                               & P2T & 76.9 & 25.4 \\
                               & \textbf{PVTv2} & \textbf{83.2} & \textbf{14.5} \\
  \bottomrule
\end{tabular}
\end{table}

\subsubsection{Ablation Study on Backbone}
To bridge the semantic gap between the encoder and decoder, existing methods typically adopt the U-shaped architecture and employ skip connections to enhance semantic consistency. In our proposed DTEA model, we design a novel skip connection module that more effectively fuses multi-scale semantic features between the encoding and decoding stages. To systematically evaluate the adaptability and robustness of the proposed module across different encoder-decoder backbones, we integrate several widely used CNNs and Transformers, including ResNet-50~\cite{27he2016deep}, Res2Net~\cite{28gao2019res2net}, ResNeSt-50~\cite{29zhang2022resnest}, ViT-B/16~\cite{30dosovitskiy2020image}, P2T~\cite{31wu2022p2t} and PVTv2~\cite{32wang2022pvt}. As shown in \cref{tab:backbone_grouped}, the proposed skip connection module exhibits robust generalization capabilities and adapts to diverse encoder-decoder architectures. In particular, PVTv2, with its strong multi-scale feature extraction capability, achieves the highest DSC of 83.2\% when combined with DTEA. Its inherent pyramid structure and efficient attention mechanisms provide the most suitable multi-scale feature foundation for our module.

\section{Conclusion}
In this study, we present DTEA, a medical image segmentation model that incorporates a novel skip connection design, consisting of STR and EPG, to bridge the semantic gap between the encoder and decoder. The model not only enhances multi-scale semantic fusion but also guides the network to focus on critical spatial regions, thereby better preserving important anatomical structures. We conducted systematic experiments on three tasks including polyp segmentation, skin lesion segmentation, and multi-organ segmentation. The results demonstrate that DTEA achieves consistently strong performance across all challenging datasets, validating its effectiveness and robust generalization capability.

\bibliographystyle{IEEEtran}
\bibliography{ref}

\end{document}